\documentclass[12pt]{article}

\usepackage{amssymb,amsmath,amsfonts,eurosym,geometry,ulem,graphicx,caption,color,setspace,sectsty,comment,footmisc,caption,pdflscape,subfigure,array}
\usepackage[draft,colorlinks=true,linkcolor=blue,citecolor=blue]{hyperref}

\usepackage[utf8]{inputenc}

\newcolumntype{L}[1]{>{\raggedright\let\newline\\arraybackslash\hspace{0pt}}m{#1}}
\newcolumntype{C}[1]{>{\centering\let\newline\\arraybackslash\hspace{0pt}}m{#1}}
\newcolumntype{R}[1]{>{\raggedleft\let\newline\\arraybackslash\hspace{0pt}}m{#1}}

\newcommand*\dd{\text{d\kern 0.03em}}
\newcommand*\dz{\text{d\kern 0.05em}z_t}
\newcommand*\dq{\text{d\kern 0.05em}z_t^{\mathbb{Q}}}

\usepackage[
    backend=biber,
    maxcitenames=2,
    style=authoryear,
    natbib=true,
    url=true, 
    uniquename=false,
    doi=true,
    eprint=false
]{biblatex}
\usepackage{tabularx}
\usepackage{caption}
\newcommand{\hi}[1]{} 

\usepackage{booktabs}

\usepackage{rotating}

\usepackage{algorithm}
\usepackage{algpseudocode}

\usepackage[super]{nth}

\usepackage{titlesec}
\titleformat{\section}{\normalfont\large\centering}{\bfseries {\bfseries\thesection.}}{0.5em}{\MakeUppercase}
\titlespacing*{\section}{0pt}{2em}{0.25em}

\titleformat{\subsection}{\normalfont\bfseries\large\centering}{\thesubsection.}{0.5em}{}
\titlespacing*{\subsection}{0pt}{1em}{0.3em}{}

\titleformat{\subsubsection}{\normalfont\large\centering}{\thesubsubsection.}{0.5em}{}
\titlespacing*{\subsubsection}{0pt}{1em}{0.3em}{}

\begin{document}

\begin{titlepage}
\title{Time Series (re)sampling using Generative Adversarial Networks\footnote{Acknowledgements: The authors gratefully acknowledge support from the Google Tensorflow Research Cloud (TFRC). PyTorch code for this paper is available on request. We also thank Giovanni Mellace and Peter Sandholt Jensen for useful comments.}}
\author{Christian M. Dahl\footnote{Department of Business and Economics, University of Southern Denmark, cmd@sam.sdu.dk} $\quad$ Emil N. Sørensen\footnote{School of Economics, University of Bristol, e.sorensen@bristol.ac.uk}}
\date{\today}
\maketitle
\begin{abstract}
\noindent We propose a novel bootstrap procedure for dependent data based on Generative Adversarial networks (GANs). We show that the dynamics of common stationary time series processes can be learned by GANs and demonstrate that GANs trained on a single sample path can be used to generate additional samples from the process. We find that temporal convolutional neural networks provide a suitable design for the generator and discriminator, and that convincing samples can be generated on the basis of a vector of iid normal noise. We demonstrate the finite sample properties of GAN sampling and the suggested bootstrap using simulations where we compare the performance to circular block bootstrapping in the case of resampling an AR(1) time series processes. We find that resampling using the GAN can outperform circular block bootstrapping in terms of empirical coverage.

\bigskip
\end{abstract}
\thispagestyle{empty}
\end{titlepage}
\pagebreak \newpage

\doublespacing

\section{Introduction}
Generative Adversarial Nets (GANs) were introduced by \citet{goodfellow2014generative}. Based on an initial training sample, GANs can learn to generate additional data that looks similar. GANs are intensely researched in the deep learning literature \citep{radford2015unsupervised,salimans2016inceptionscore,gulrajani2017improved,arjovsky2017wasserstein} but have received minor attention in time series analysis. There are examples of GANs being explored for structural models \citep{kaji2018deepinf} and estimation of treatment effects \citep{athey2019using}. However, these are both in the cross-sectional iid setting. \citet{esteban2017real} suggest that GANs can be used for what they call ``medical'' time series but they lack a clear definition of the data generating process (DGP) and correspondingly measures of quality for the learned model. Recently, \citet{wiese2020quantgan} described a GAN for financial time series that can reproduce some of the stylised facts of such series using a temporal convolution architecture related to the one suggested in this work. \citet{smith2020conditional} also outlined a method for training GANs on time series using spectrograms. Unlike \citet{wiese2020quantgan,smith2020conditional} our focus is on the general applicability of GANs as a bootstrap method for dependent processes. The potential usefulness of GANs for such time series bootstraps is briefly mentioned in recent work by \citet{haas2020statistical}.

GANs have frequently been applied for image synthesis \citep{goodfellow2014generative,radford2015unsupervised,arjovsky2017wasserstein} and the generated samples are often evaluated using measures such as Inception Score \citep{salimans2016inceptionscore} or Frèchet Inception Distance \citep{heusel2017fid}. Both measures utilise a neural network trained for image recognition to attempt to assess the visual quality of the generated samples. However, these are heuristics and it is difficult to construct a theoretically motivated notion of quality. Contrary to \citet{esteban2017real}, we argue that it is straightforward to assess the basic properties of the generated samples in a time series context as the theoretical properties of many time series are well understood -- contingent on considering explicitly defined data generating processes. A similar point is made by \citet{wiese2020quantgan}. 

We show that stationary autoregressive time series processes -- exemplified by the AR(1) process -- can be learned by GANs trained on a single sample path and find that temporal convolutional neural networks provide a suitable design for the generator and discriminator. 

Bootstrapping dependent data -- such as samples from a time series process -- has received long-standing attention in the literature, see e.g. the overview of block bootstrapping by \citet{lahiri1999theoretical} and the newer contributions of \citet{politis2001tapered,shao2010wild}. We suggest that GANs provide a novel approach to such bootstrapping which we call the generative bootstrap (GB). The theoretical properties of GANs -- and hence our suggested generative bootstrap -- is an active area of research, see \citep{biau2018theoretical,biau2020theoretical,haas2020statistical} and we instead contribute by analysing the finite sample properties of generative bootstrapping using simulations where we compare the performance against Circular Block Bootstrapping (CBB) \citep{politis1992circular} for the AR(1) process. In particular, we recover the parameters of the true data generating process using the simulated data and thereby show that the generative model has learned at least a minimum of characteristics of the process. Further, we find that resampling using the generative model can outperform CBB for dependent data on the empirical coverage of percentile confidence intervals.

Our main contributions are: (1) we show that GANs can learn the dynamics of stationary autoregressive time series processes, (2) we find that temporal convolutions provide a working architecture for the discriminator and generator networks, and (3) we show that GANs can be used to resample from dependent data with suggestive finite-sample improvements in empirical coverage over CBB. (1) and (2) are also discussed by \citet{wiese2020quantgan}. However, \citet{wiese2020quantgan} do not consider the more general applicability of GANs to time series bootstrapping which we define and subsequently evaluate in our simulations.

In Section~\ref{sec:gan} we review two common GANs and discuss how they can be trained to generate samples from a time series based on an initial sample path. Section~\ref{sec:genbootstrap} discusses an algorithm for using the trained GAN to bootstrap dependent data. In Section~\ref{sec:simulations} we provide simulation evidence of the quality of the learned GAN and the performance when it is used for bootstrapping. Section~\ref{sec:conclusion} concludes.

\section{Generative adversarial nets}\label{sec:gan}
\subsection{Basic GAN} 

The concept of GANs can be introduced intuitively as follows. Assume that a real sample of data is drawn from the unknown distribution $F_X$ and assume that we have another arbitrary, but known, distribution $F_Z$. The generator $G$ is a function that transforms a sample from $F_Z$ into a sample that looks like it is drawn from the real data distribution $F_X$. The discriminator $D$ is a function that tries to determine if a given sample is drawn from the real data distribution $F_X$ or not. The two models are set to play a game against each other. The generator tries to fool the discriminator by generating fake samples that look as real as possible, and the discriminator tries to detect the generator's forgery by determining if it got a real or fake sample.

Let $G$ and $D$ be specified up to the finite dimensional parameters $\theta_G$ and $\theta_D$ respectively. Also, let $x_{real}$ denote some generic real sample from distribution $F_X$ and $x_{fake} = G(z; \theta_G), z \sim F_Z$ a generated sample. 

\citet{goodfellow2014generative} suggest solving the minimax problem
\begin{equation}
\min_{\theta_G}\max_{\theta_D} \; \mathbb{E}_{F_X} \! \log D(x; \theta_D) + \mathbb{E}_{F_Z} \! \log(1 - D(G(z; \theta_G); \theta_D)).
\label{eq:basicgan}
\end{equation}
In practice \citet{goodfellow2014generative} separate the minimisation and maximisation steps into
\begin{align*}
&\max_{\theta_D} \; \mathbb{E}_{F_X} \! \log D(x; \theta_D) + \mathbb{E}_{F_Z} \! \log(1 - D(G(z; \theta_G); \theta_D)) \\
&\min_{\theta_G} \; \mathbb{E}_{F_Z} \! \log(1 - D(G(z; \theta_G); \theta_D))
\end{align*}
and iterate between these to learn the discriminator and generator using batching and stochastic gradient descent. We skip the details here, but describe the training in detail for the Wasserstein GAN in the following section. Algorithm~\ref{algo:gantrain} provides an overview of the training algorithm.

\citet{biau2018theoretical} and \citet{goodfellow2014generative} argue that, under a set of assumptions, the optimal discriminator in the minimax formulation in Equation~\ref{eq:basicgan} is related to the Jensen-Shannon divergence between the distributions of the real and generated data. If $F_G$ is the distribution of the transformation $G(Z; \theta_G), Z \sim F_Z$ and the optimal discriminator is in the class of functions $\{ D(\cdot, \theta_D) : \theta_D \in \Theta \}$ where $\Theta$ is some parameter space then the solution to the \emph{maximisation} problem in Equation~\ref{eq:basicgan} is the Jensen-Shannon divergence $JS$ between $F_X$ and $F_G$ \citep{biau2018theoretical,goodfellow2014generative},
\begin{equation}
\max_{\theta_D \in \Theta} \; \mathbb{E}_{F_X} \! \log D(x; \theta_D) + \mathbb{E}_{F_Z} \! \log(1 - D(G(z; \theta_G); \theta_D)) =  2 \text{JS}(F_X, F_G) - \log 4 \label{eq:js}
\end{equation}
and, heuristically, if we assume the discriminator is optimal then the generator is solving the problem
\begin{equation}
\min_{\theta_G} \text{JS}(F_X, F_G). \label{eq:infgan}
\end{equation}
As noted by \citet{biau2018theoretical}, this seems to have motivated work on investigating other divergences/distances in the context of GANs. One such distance is the Wasserstein distance which we will consider in the following section.

\begin{algorithm}\setstretch{1.1}
\begin{algorithmic} 
\For {$i = 1, 2, ..., N$}:
	\State{$z \gets \text{Sample}(F_Z)$} \Comment{\textbf{Discriminator update}}
	\State{$x_{fake} \gets G(z; \theta_G)$}
	\State{$x_{real} \gets \text{Sample}(\mathcal{X})$} \Comment{$\mathcal{X}$ is a given collection of samples from $F_X$}
	\State{$L_D^{(b)} \gets \frac{1}{n_b} \sum_i \log D(x_{real}; \theta_D) + \log\left( 1 - D(x_{fake}; \theta_D)\right) $} \Comment{Discriminator loss}
	\State{$\theta_D \gets \theta_D - lr_D \nabla_{\theta_D} \! L_D^{(b)}$} \Comment{Parameter update}
\item
	\State{$z \gets \text{Sample}(F_Z)$} \Comment{\textbf{Generator update}}
	\State{$x_{fake} \gets G(Z; \theta_G)$}
	\State{$L_G^{(b)} \gets \frac{1}{n_b} \sum_i \log(1 - D(x_{fake}; \theta_D))$} \Comment{Generator loss}
	\State{$\theta_G \gets \theta_G - lr_G \nabla_{\theta_G} \! L_G^{(b)}$} \Comment{Parameter update}
\EndFor
\end{algorithmic}
\caption{\strut GAN \citep{goodfellow2014generative}}
\label{algo:gantrain}
\end{algorithm}

\subsection{Wasserstein GAN}
\citet{arjovsky2017wasserstein} argue that an alternative distance measure in Equation~\ref{eq:infgan} is the order-1 Wasserstein (Earth-Mover) distance which results in the Wasserstein GAN. Consider informally two probability measures $\mathbb{P}$ and $\mathbb{Q}$ defined on a suitable common probability space $(\mathcal{M}, \cdot)$. The Wasserstein distance $W_1$ between $\mathbb{P}$ and $\mathbb{Q}$ is defined as
\begin{equation}
W_1(\mathbb{P}, \mathbb{Q}) = \inf_{\mathbb{V} \in \Pi} \mathbb{E}_{\mathbb{V}}{||x - y||} \label{eq:wdistance}
\end{equation}
where, with abuse of notation, $\Pi$ is the set of all joint probability measures $\mathbb{V}(x, y)$ with marginal probabilities $\mathbb{P}(x)$ and $\mathbb{Q}(y)$, and $|| \cdot ||$ is the absolute value norm \citep{arjovsky2017wasserstein}. Here $\mathbb{E}_{\mathbb{V}}$ denotes expectation under the probability measure $\mathbb{V}$. 
\citet{arjovsky2017wasserstein} argue that Equation~\ref{eq:wdistance} is equivalent to 
\begin{equation}
W_1(\mathbb{P}, \mathbb{Q}) = \sup_{D \in \mathcal{F}} \mathbb{E}_{\mathbb{P}} \, D(x) - \mathbb{E}_{\mathbb{Q}} \, D(x)
\label{eq:krd}
\end{equation}
where $\mathcal{F}$ is the set of real-valued Lipschitz functions on $\mathcal{M}$ with Lipschitz constant 1. 
In Equation~\ref{eq:krd} we have conveniently denoted the function to be optimised over by $D$ as we can consider it to play the role of a discriminator. Given the discriminator, the generator would like to minimise the distance between the generated data and real data, if the laws of generated and real data are given by $\mathbb{P}$ and $\mathbb{Q}$ then the generator is solving the problem $\inf_{G} W_1(\mathbb{P}, \mathbb{Q})$. This is analogous to Equation~\ref{eq:infgan} but the Jensen-Shannon divergence $JS$ has been replaced by the Wasserstein distance $W_1$.
A primary issue in operationalising Equation~\ref{eq:krd} is enforcing the Lipschitz condition on $D$. For example, say we learn $D$ using a neural network, how do we constrain this network to only learn Lipschitz-1 functions? Let $G$ and $D$ be specified up to the finite dimensional parameters $\theta_G$ and $\theta_D$ respectively. Also, let $x_{real}$ denote some generic real sample from distribution $F_X$ and $x_{fake} = G(z; \theta_G), z \sim F_Z$ a generated sample. Now based on Equation~\ref{eq:krd} \citet{arjovsky2017wasserstein} suggest solving the minimax problem

\begin{equation}
\min_{\theta_G} \max_{\theta_D} \; \mathbb{E}_{F_X} D(x; \theta_D) - \mathbb{E}_{F_Z} D(G(z; \theta_G); \theta_D) \label{eq:wganmainproblem}
\end{equation}
subject to $D(\, \cdot \,; \theta_D) \in \mathcal{F}$.
A simple training algorithm for solving the problem (\ref{eq:wganmainproblem}) would be splitting it into a min and max step, and iterate between them \citep{goodfellow2014generative}
\begin{align}
&\max_{\theta_D} \; \mathbb{E}_{F_X} D(x; \theta_D) - \mathbb{E}_{F_Z} D(G(z; \theta_G); \theta_D) \label{eq:maxdis}\\
&\min_{\theta_G} \; - \mathbb{E}_{F_Z} D(G(z; \theta_G); \theta_D). \label{eq:mingen}
\end{align}

\citet{gulrajani2017improved} recognise that a function is Lipschitz-1 if and only if the norm of the gradient is 1 or less everywhere, so they suggest a gradient penalty to enforce the Lipschitz condition in the discriminator
$$P(\theta_D) = \mathbb{E}_{F_{\tilde{X}}} \left( |\!| \nabla_{\tilde{x}} D(\tilde{x}; \theta_D) |\!|_2 - 1\right)^2$$
where $|| \cdot ||_2$ is the $l_2$ norm. Note that $\tilde{x} = a x_{real} + (1-a) x_{fake}$ is a convex combination of $x_{real}$ and $x_{fake}$ with uniform random weight $a \sim \text{U}(0, 1)$, and we let $F_{\tilde{x}}$ denote the distribution of these convex combinations. This procedure is motivated heuristically in \citet{gulrajani2017improved} and is a less computationally intensive way of enforcing the Lipschitz constraint across all possible $x$. Under the gradient penalty the discriminator objective function is now
\begin{equation}
\max_{\theta_D} \; \mathbb{E}_{F_X} D(x; \theta_D) - \mathbb{E}_{F_Z} D(G(z; \theta_G); \theta_D) + \lambda P(\theta_D) \label{eq:penalizedwgandis}
\end{equation}
where the weight of the gradient penalty is adjusted by the hyper parameter $\lambda$.

Let $\{(z_i, x_{i,real})\}_{i=1}^{n_b}$ constitute a (mini) batch of data where $z_i$ is noise sampled from $F_Z$ and $x_{i,real}$ is a real sample. During training we minimise the batch discriminator loss
\begin{align*}
L^{(b)}_D &= \frac{1}{n_b} \sum_{i=1}^{n_b} D(x_{i,fake}; \theta_D) - D(x_{i,real}; \theta_D) + \lambda \frac{1}{n_b} \sum_{i=1}^{n_b} \left( |\!| \nabla_{\tilde{x}} D(\tilde{x}_i; \theta_D) |\!|_2 - 1\right)^2, \tag{D1} \label{eq:wgandis}\\
\tilde{x}_i &= a x_{i,real} + (1-a) x_{i,fake}
\end{align*}
which is the empirical and batched equivalent of Equation~\ref{eq:penalizedwgandis}, and recall that $x_{i,fake} = G(z_i; \theta_G)$. As per usual, the batch gradients $\nabla_{\theta_D} L_{D}^{(b)}$ serve as unbiased estimates of $\nabla_{\theta_D} L_D$ (i.e. here $L_D$ is the loss over the entire training sample while $L_D^{(b)}$ is the loss in the batch only, so for $L_D$ the sums run over $(1, ..., n)$ instead of $(1, ..., n_b)$) which allow us to do stochastic gradient descent on the parameters $(\theta_D, \theta_G)$. The first sum in (\ref{eq:wgandis}) amounts to the discriminator objective in \citet{arjovsky2017wasserstein} while the second sum corresponds to the gradient penalty suggested by \citet{gulrajani2017improved}. 

Similarly, for the generator we minimise the batch generator loss
\begin{align*}
L^{(b)}_G &= \frac{1}{n_b} \sum_{i=1}^{n_b} - D(G(z; \theta_G); \theta_D) \tag{G1}\label{eq:wgangen}
\end{align*}
corresponding to Equation~\ref{eq:mingen}.  By alternating between the objectives (D1) and (G1) we can learn the parameters of $G$ and $D$. The complete training algorithm of \citet{gulrajani2017improved} is given in Algorithm~\ref{algo:wgantrain} in pseudo-code. Notice that during the discriminator updates the gradient is with respect to $\theta_D$ and in the generator updates with respect to $\theta_G$. Algorithm~\ref{algo:wgantrain} uses Stochastic Gradient Descent (SGD) to update the parameters, but more sophisticated optimisation methods could also be applied, e.g. ADAM \citep{kingma2014adam}.

\begin{algorithm}\setstretch{1.1}
\begin{algorithmic}[1]
\For {$i = 1, 2, ..., N$}:
\For {$j = 1, 2, ... N_{discriminator}$}: \Comment{\textbf{Discriminator updates}.}
	\State{$z \gets \text{Sample}(F_Z)$} \label{alg:line:samplenoise}
	\State{$x_{fake} \gets G(z; \theta_G)$} \label{alg:line:generatefake}
	\State{$x_{real} \gets \text{Sample}(\mathcal{X})$} \label{alg:line:sample} \Comment{$\mathcal{X}$ is a fixed collection of samples from $F_X$}
	\item
	\State{$\tilde{x} \gets a x_{real} + (1-a) x_{fake}$} \Comment{$a$ is drawn from the uniform distribution $\text{U}(0,1)$.}
	\State{$P^{(b)} \gets \frac{1}{n_b} \sum_{b=1}^{n_b} \left( || \nabla_{\!\! \tilde{x}} D(\tilde{x}; \theta_D) ||_2 - 1 \right)^2$} \Comment{Gradient penalty.}
	\State{$L_D^{(b)} \gets \frac{1}{n_b} \sum_{b=1}^{n_b} D(x_{fake}; \theta_D) - D(x_{real}; \theta_D) + \lambda P^{(b)}$} \Comment{Discriminator loss.}
	\State{$\theta_D \gets \theta_D - lr_D \, \nabla_{\!\!\theta_D} L_D^{(b)}$} \Comment{Parameter update, $lr_D$ is the learning rate.}
\EndFor
\item
\For {$j = 1, 2, ... N_{generator}$} \Comment{\textbf{Generator updates}.}
	\State{$z \gets \text{Sample}(F_Z)$}
	\State{$x_{fake} \gets G(z; \theta_G)$}
	\item
	\State{$L_G^{(b)} \gets \frac{1}{n_b} \sum_{b=1}^{n_b} -D(x_{fake}; \theta_D)$} \Comment{Generator loss.}
	\State{$\theta_G \gets \theta_G - lr_G \, \nabla_{\!\!\theta_G} L_G^{(b)}$} \Comment{Parameter update, $lr_G$ is the learning rate.}
\EndFor
\EndFor
\end{algorithmic}
\caption{\strut Wasserstein GAN with Gradient Penalty \citep{arjovsky2017wasserstein,gulrajani2017improved}.}
\label{algo:wgantrain}
\end{algorithm}

The GAN formulation above does not necessarily impose how we should parameterise the discriminator $D$ and generator $G$. However, in practice, they are commonly learned using neural networks with exact parameterisations depending on the application.\footnote{For example, \citet{radford2015unsupervised} use convolutions for images while \citep{athey2019using} use fully-connected layers in the context of treatment effects.} \citet{hornik1990universal} showed that neural networks with fully-connected layers enjoy universal approximation properties and hence are a natural choice. We do not give an introduction to neural networks and their terminology but refer to the textbook treatment by \citet{goodfellow2016deep}. 

Consider a time series process $Y = \{ Y_t : t \in \mathcal{T} \}$ indexed by time $t$. A time series has the defining property that information flow is unidirectional, so the state of the process at time $t$, $Y_t$, can only depend on past information $(Y_{t-1}, Y_{t-2}, ...)$ while the future is unknown. This constraint is useful when we choose the parameterisation of $G$ and $D$. 

The GAN in \citet{esteban2017real} relied on recurrent neural networks (RNNs) to model time series. We pursue a different approach and base the generator and discriminator on stacked dilated temporal convolutions (DTC) used by \citet{oord2016wavenet} for audio generation. We will refer to this as the TC-architecture. The temporal convolutions are similar to conventional convolutions -- see \citep[Chp. 3]{goodfellow2016deep} -- but they enforce the unidirectional flow of information. They were applied to time series forecasting by \citet{borovykh2017conditional} and \citet{sen2019think}. In particular, \citet{borovykh2017conditional} showed that DTC networks outperform RNNs in several forecasting problems and are easier to train even for long-range dependence. Very recently, \citet{wiese2020quantgan} similarly suggested temporal convolutions based on \citep{oord2016wavenet} for financial time series modelling with GANs. 
The dilation of the temporal convolutions increases the receptive field -- in context of time series this is the number of lags that the model can accommodate at once -- while limiting the number of parameters \citep{oord2016wavenet,borovykh2017conditional}.

\begin{figure}\centering
\includegraphics[width=0.3\textwidth]{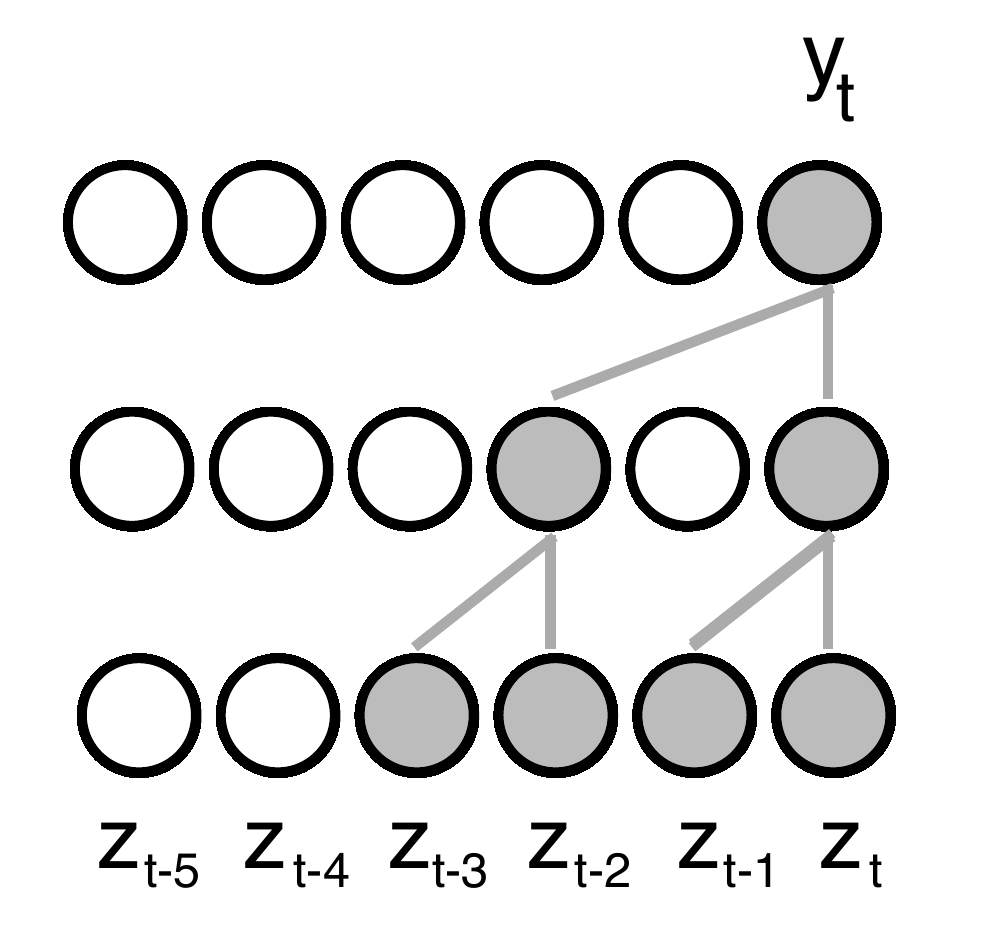}
\caption{Adapted from Figure~2 in \citep{oord2016wavenet}. An illustration of the DTC network. The output $y_t$ at time $t$ is a function of the present and past noise terms $(z_t, z_{t-1}, z_{t-2}, z_{t-3})$. We generate the output samples as we slide across the noise terms.}
\label{fig:dtc}
\end{figure}

In practice, temporal convolutions can be implemented as conventional one-dimensional convolutions with appropriate zero-padding applied to the input, see \citep{oord2016wavenet}. If we stack $d$ DTC layers with kernel size $2$ where the dilation for layer $i$ is $2^i$ then the total receptive field size at the final layer will be \citep{yu2015multiscale}
\begin{equation}
p =  \sum_{i=1}^d 2^i = 2^{d+1}-1. \label{eq:tcpadding}
\end{equation}
For illustration, assume that our generator $G$ consists of $d$ DTC layers. To generate a time series of length $b$ we slide the DTC layers over a sequence of $(b + p)$ iid noise terms
$$(z_1, z_2, ..., z_{p+b}), \quad z_t \sim F_Z$$
where $F_Z$ is some arbitrarily chosen distribution and $p$ is the receptive field size given in Equation~\ref{eq:tcpadding}. During the GAN training, the parameters in the DTC layers learn to transform this sequence of iid noise into observations from the time series. This is illustrated in Figure~\ref{fig:dtc} for a generator with two DTC layers. On the other hand, the discriminator $D$ considers sequences of observations from the generated or real time series $(y_1, y_2, ..., y_T)$ and learns the parameters in the DTC layers to distinguish between real and generated samples. We will detail this process in the context of bootstrapping in Section~\ref{sec:genbootstrap}, and provide a complete example of the architecture in Section~\ref{sec:simulations}.

\section{Generative bootstrap}\label{sec:genbootstrap}
We propose the GAN with temporal convolution layers as a method to resample from a time series process. This procedure is called the Generative Bootstrap (GB). The GB is composed of two stages: (1) the GAN is trained on an initial sample from the true DGP using blocking -- the \emph{training stage}. (2) samples are generated from the generator of the trained GAN -- the \emph{sampling stage}.

The initial sample from the true DGP is re-sampled using a moving block scheme similar to the Moving Block Bootstrap (MBB) \citep{kunsch1989jackknife,liu1992moving} and $n_b$ of such blocks constitute a batch of data that are used to perform one iteration over the batch losses in Equation~\ref{eq:wgandis}-\ref{eq:wgangen}. Many iterations are performed until the GAN losses stabilise. For sampling, the discriminator is discarded and we feed noise into the generator from an arbitrary distribution $F_Z$. The generator output is used as a sample to calculate one set of bootstrap statistics. This procedure is repeated for an arbitrary number of samples and the collection of statistics is used to form the GB estimates.

We will see that this differs from conventional block bootstrapping in two respects. Firstly, a conventional MBB would sample blocks of size $b$ with replacement from the initial sample and stack these into one sample path matching the length of the initial sample. This stacked sample is then used to calculate bootstrap statistics. In the GB these blocks are not stacked, but fed as individual samples to train the GAN. Secondly, sampling the GAN does not have to use stacking and the GAN can provide a sample of any size once it has been trained. The size of the sample is determined by the noise terms supplied to the generator. 

We proceed to discuss the training and sampling stages below. Algorithm~\ref{algo:gb} provides an overview.

\begin{algorithm}\setstretch{1.1}
\begin{algorithmic}[1]
\For {$r = 1, 2, ..., R$}: \Comment{Training stage}
\State Use Algorithm~\ref{algo:wgantrain} but batches consist of resampled blocks with size $b_1$.
\EndFor
\item 
\For {$s = 1, 2, ..., S$}: \Comment{Sampling stage}
\State $x \gets \text{Sample}(model, b_2)$
\State $estimates[s] \gets \text{Statistic}(x)$
\EndFor
\end{algorithmic}
\caption{\strut Generative Bootstrap}
\label{algo:gb}
\end{algorithm}

\textbf{Traning stage}. The training stage mimics Algorithm~\ref{algo:wgantrain} but uses moving blocks to re-sample the initial sample, see Line~\ref{alg:line:sample} in Algorithm~\ref{algo:wgantrain}. Let $y^*_i = (y^*_{1,i}, y^*_{2,i}, ..., y^*_{T,i})$ be the initial sample from the true DGP. We perform a blocking procedure identical to the moving block bootstrap. Define each of the $(T - b_1 - 1)$ overlapping blocks by $B^*_j = (y_{1j} = y^*_{1+j,i}, ..., y_{2j} = y^*_{1+j+b_1,i})$ where the block size is $b_1 < T$. A batch of training data is given by randomly sampling $n_b$ blocks from $\{ B^*_1, B^*_2, ..., B^*_{T - b_1 - 1} \}$ without replacement, denote this batch of true blocks by $\mathcal{Y}^*$. 

Next, we sample the generator noise (see Line~\ref{alg:line:samplenoise} of Algorithm~\ref{algo:wgantrain}) from a multivariate standard normal distribution with an identity variance-covariance matrix, but the distribution $F_Z$ could be selected arbitrarily. To generate a sample path of length $b_1$ we need $(b_1 + p)$ noise terms where $p$ is the receptive field in the DTC layers -- see Equation~\ref{eq:tcpadding}. The noise terms are used to generate a block sample $B_{j} = (G(z_1, ..., z_p; \theta_G), G(z_{1+1}, ..., z_{p+1}; \theta_G), ..., \allowbreak G(z_{b_1}, ..., z_{b_1+p}; \theta_G))$, see Line~\ref{alg:line:generatefake} of Algorithm~\ref{algo:wgantrain}, and $n_b$ of such blocks are generated to form a batch of fake blocks, denote them by $\mathcal{Y}$. The true and fake samples are fed to the discriminator and it tries to distinguish between them. This requires that the blocks in $\mathcal{Y}^*$ and $\mathcal{Y}$ have the same size. This procedure of sampling true and fake blocks and feeding them to the discriminator constitutes the training stage. In practice, we iterate over Equation~\ref{eq:wgandis}-\ref{eq:wgangen} until the losses stabilise. Equation~\ref{eq:wgandis} requires both a fake and true batch per iteration, while Equation~\ref{eq:wgangen} needs only a fake batch.

\textbf{Sampling stage}. Let $G(\, \cdot \,; \hat\theta_G)$ be the learned generator from the training stage. Once trained, the generator should produce samples mimicking the true DGP that generated the initial sample $y_i^*$. To generate a sample of length $b_2$, we first sample a sequence of noise vectors from $F_Z$
$$z_i = (z_{1,i}, z_{2,i}, ..., z_{b_2 + p,i}), \quad z_{t,i} \sim F_Z$$
where $p$ is again given in Equation~\ref{eq:tcpadding}. As in the training stage, $F_Z$ is a multivariate standard normal distribution with an identity variance-covariance matrix. Any distribution could be used, the important point is that the training and sampling stages use the \emph{same} distribution for $F_Z$. 
Next we obtain a generated sample path $y_i$ by passing the noise vectors through the learned generator,
$$y_i = (G(z_{1,i}, ..., z_{p,i}; \hat\theta_G), G(z_{1+1,i}, ..., z_{p+1,i}; \hat\theta_G), ..., \allowbreak G(z_{b,i}, ..., z_{b_2+p,i}; \hat\theta_G)).$$
A single sequence of innovation vectors $z_i = (z_{1,i}, ..., z_{b_2 +p,i})$ generates one sample path $y_i$ of length $b_2$. We can repeat this process to obtain an arbitrary number of sample paths. 

Under the proposed TC-architecture the generator can sample a block of any length from the underlying process and hence we are not restricted to the block size on which the model was trained, i.e. it is perfectly acceptable if $b_1 \not= b_2$. This does not necessarily hold for all choices of architecture, e.g. a fully-connected network would not have this property. This is a very attractive feature of the TC and GAN approach as it alleviates the need to stack individual blocks in a way that might break the dependence structure of the time series. We can simple choose the sampling block size to be equal to the size of the initial sample path, so $b_2 = T$. 

When $b_2 < T$ we refer to it as \emph{blocked sampling}, while $b_2 = T$ is called \emph{complete sampling}.

\textbf{Bootstrap statistics}. Let $G(\cdot; \hat\theta_G)$ be the learned generator from the training stage that has been trained on a single initial sample $y_i^*$ from the true DGP. We now discuss how to calculate bootstrap statistics on the GAN samples. Assume that we are interested in parameter $\phi$ which has a suitable estimator $\hat\phi$. 
We use the sampling procedure from the previous section to obtain $m$ samples from the learned generator $G(\, \cdot \, ; \hat\theta_G)$, denote these samples by $(y_1, y_2, ..., y_m)$ where $y_i = (y_{1,i}, ..., y_{T,i}), i = 1, ..., m$. Each $y_i$ is considered a realisation of the DGP that produced the initial training sample for the GAN. We calculate the bootstrap statistics $\hat\phi_{i} \equiv \hat\phi(y_i), \quad i = 1, ..., m$ resulting in $m$ estimates $(\hat\phi_{1}, \hat\phi_{2}, ..., \hat\phi_{m})$. Similar to conventional bootstrapping \citep{efron1981percentile}, the GB variance estimate of $\hat\phi$ is
\begin{equation}
\hat{\sigma}_{GB,\hat\phi} = \frac{1}{m} \sum_i (\hat\phi_{i} - \hat\phi_{GB})^2.
\end{equation}
The $(1- \alpha)$ GB confidence interval (CI) for $\phi$ is the $(1-\alpha)$ percentile CI \citep{efron1981percentile} constructed using the empirical quantiles\footnote{A possible improvement on this confidence interval would be the bias-correction techniques outlined in \citep{efron1987bca}.} $(\alpha/2, 1 - \alpha/2)$ of $(\hat{\phi}_{i})_i$,
\begin{equation}
\hat{I}_{\phi,1-\alpha} = \left[ \hat{\phi}_{(\alpha/2)}, \, \hat{\phi}_{(1-\alpha/2)} \right].
\end{equation}

\section{Simulations}\label{sec:simulations}
In this section we will illustrate the performance of the GB by using simulations and by making comparisons to the established CBB approach for bootstrapping dependent processes. For simplicity of exposition we base our illustrations on the AR(1) as the data generating process.   
\subsection{AR(1) process}
The simulation design is as follows. The true DGP is a zero mean and stable AR(1) process
\begin{equation}
y_{t} = \phi \, y_{t-1} + \epsilon_t, \quad \epsilon_t \sim \text{N}(0, 1)
\label{eq:ar1model}
\end{equation}
with $\phi = 0.5, 0.8, 0.9$. For each replication, a sample path of length $T = 1,000$ is generated from Equation~\ref{eq:ar1model}. This sample is used to train the GAN with a training block size of $b_1 = 150$ and batch size $n_b = 64$. Once training is complete, we sample $10,000$ sample paths from the GAN. These samples are used for two purposes:

(a) we compare the samples generated by the proposed GAN to the known properties of the DGP under complete sampling $b_2 = 1,000$. We use the generated samples to estimate the autocorrelation (ACF) and partial autocorrelation (PACF) functions over $1,000$ replications. 

(b) we compare GB and the CBB for confidence interval estimation, i.e., empirical coverage, of the least-squares estimator $\hat{\phi}_{LS}$ of $\phi$. The GB is run for $1,000$ replications and the CBB is run for $5,000$ replications. The CBB resamples from the same initial sample as is used to train the GB. We consider CBB block sizes $b_1 = 50,100,150$. The GB training block size equals 150. The number of replications for GB is lower as the simulation time is considerably higher than for CBB. A GB replication takes around 20-30 minutes while it is less than a minute for CBB. It is important to note that, in both the CBB and GB, we do not specify the dynamics of the true DGP. The GB assumes that the dynamics can be approximated by some functions of the noise vectors but these functions are not fully specified. 

The following section details the hyper parameters and training of the GAN. The two succeeding sections discuss the simulation results -- first the correlation structure of the generated samples and secondly the higher-level statistics in a bootstrapping context. 

\textbf{GAN implementation details.} We discuss the hyper parameters and network design of the GAN.

The discriminator has 6 temporal convolution layers with common kernel size $2$ and dilations $(1, 2, 4, 8, 16)$. The filters are $(8, 16, 32, 32, 64, 64)$. The output from temporal layers number 1, 2, and 6 are run through adaptive max pooling (AMP) with feature size $16$ and concatenated into a feature vector of size $48$. This is followed by two fully connected layers that regress into a single output unit. All layers use leaky ReLU activation \citep{maas2013rectifier} except for the final layer which has no activation function. The leaky ReLU avoids the zero-gradient problem of conventional ReLUs during training \citep{maas2013rectifier}. The generator has 6 temporal convolution layers that directly outputs a sample path. The filters are $(128, 64, 32, 32, 16, 1)$. Except for the last layer, all layers use the $\text{Tan}$ activation function as it -- unlike ReLU -- is symmetric. The total number of (trainable) discriminator parameters is $233,609$ while the generator has $89,921$ (trainable) parameters.

The generator noise is sampled from a multivariate standard normal distribution with an identity variance-covariance matrix. To generate a sample path of length $b$ we need $(b + p)$ noise terms where $p$ is the receptive field size in the DTC layers -- see Equation~\ref{eq:tcpadding} -- and $b$ is either $b_1$ or $b_2$ corresponding to training or sampling stage. The dimension of the noise term is a hyper parameter and can be chosen arbitrarily, in our simulations we use $256$. If we stack all the noise terms needed to produce a sample path of size $b$ then we obtain a $(b + p) \times 256$ matrix with iid standard normal distributed entries.

The GAN is trained for 5,000 steps based on a single sample from the DGP. We do not employ any (early) stopping criterion, so the GAN is always trained till the final step. The training involves iteratively minimising the batch losses, see Equation~\ref{eq:wgandis}-\ref{eq:wgangen}. Instead of stochastic gradient descent, we use the more complex Adam algorithm as it can accelerate training, see \citep{kingma2014adam}. Table~\ref{table:hypers} in the Appendix lists all hyper parameters for the GAN in this paper.

\begin{figure}[t!]
\centering
\hspace{-1em}\includegraphics[width=0.85\textwidth]{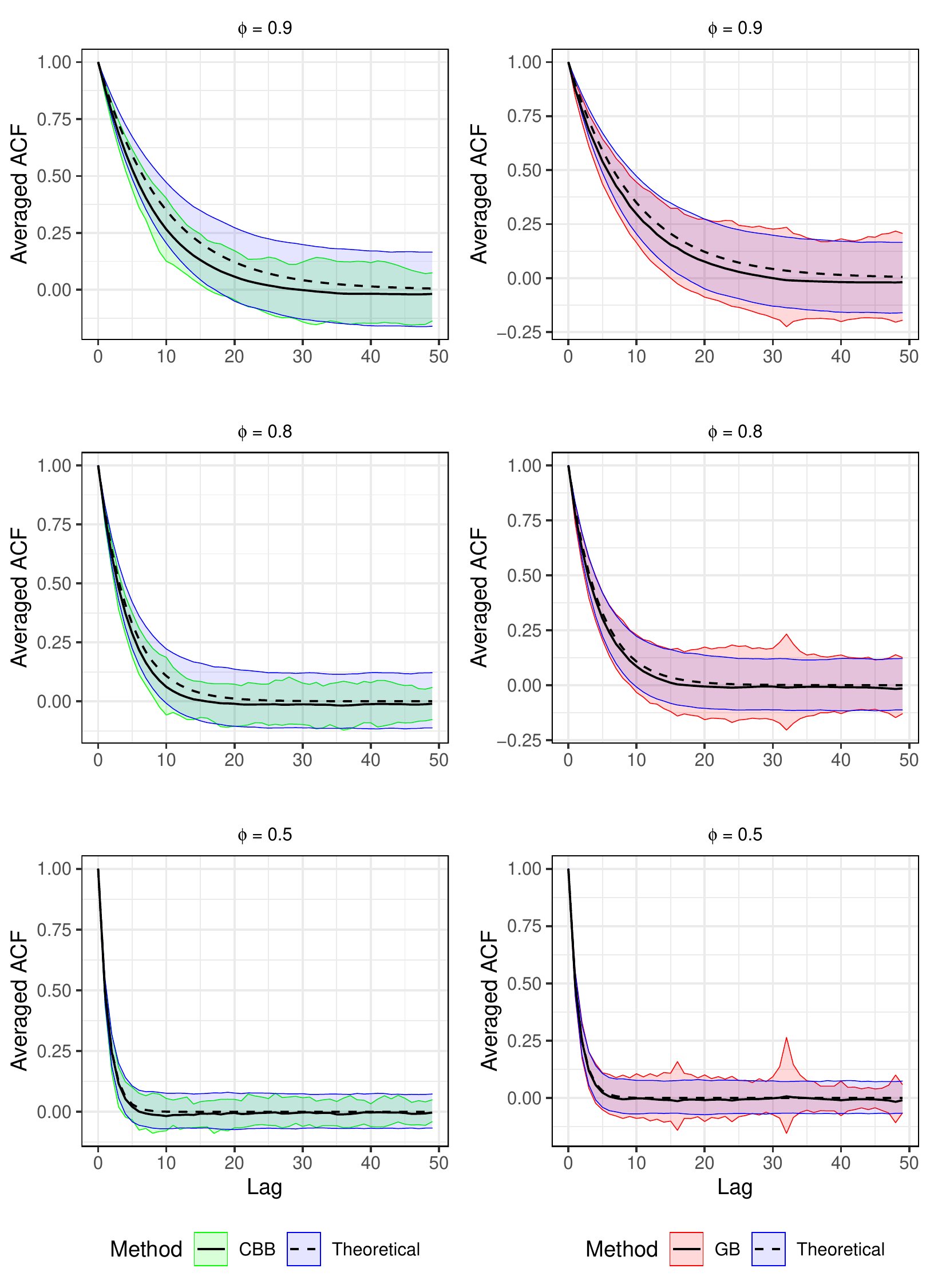}
\captionof{figure}{Theoretical (dashed line, blue ribbon) and sample autocorrelation functions for CBB (green ribbon) and GAN (red ribbon) resamples when $\phi = 0.5, 0.8, 0.9$. Block size for CBB is $150$ and training block size for GAN is $150$. The ACF estimates and confidence bands are based on $1,000$ replications of the generative model with $10,000$ samples per replication.
}
\label{figure:acf}
\end{figure}

\textbf{(a) ACF and PACF properties of the GAN samples.} We compare the samples produced by the GAN and CBB against the theoretical properties of the AR(1) process.\footnote{For the implementation of the CBB we have used the Python library by \citet{Sheppard2020}.} 
A common discussion is whether the GAN has learned to produce new samples or if it simply reproduces the original samples perfectly. If the generative model learned to perfectly replicate the original sample then the method would perform approximately on-par with CBB. Favourable bootstrapping characteristics of the GAN relative to CBB could indicate that the GAN has an advantage in capturing the dynamics of the DGP and that it does not simply replicate blocks of the original sample.

The theoretical autocorrelation function (ACF) for an AR(1) process is given by $\gamma(j) \equiv \text{Cor}(y_t, y_{t-j}) = \phi^j$ for $\phi = 0.5, 0.8, 0.9$. We estimate the ACF using generated samples under the complete sampling scheme. The ACF estimates are averaged over $1,000$ replications. Figure~\ref{figure:acf} plots the estimated ACF (full line) against the theoretical ACF (dashed line). In Figure~\ref{figure:acf} we have also included the interquartile range (IQR) for the theoretical ACF and for the ACF estimated across the $1,000$ replications. If the GAN has learned the dynamics of the AR(1) process, then the estimated and the theoretical ACF should be similar and the theoretical and the estimated IQR should be overlapping. Clearly, for higher values of the autoregressive parameter $\phi$ the persistency of the process is stronger and challenges the GAN to learn longer range dependencies.

From Figure~\ref{figure:acf} the estimated ACFs are close to their theoretical counterparts for all lags when $\phi = 0.5$. For $\phi = 0.8$ there is a small upwards bias in the estimated ACFs that is larger for the CBB, particularly, for the intermediate range of lag lengths, i.e, lags 5-20.  For $\phi = 0.9$ there is a small but noticeable bias in the estimated ACFs for all lags considered for both CB and CBB. The bias is again uniformly larger for the CBB. Importantly, all theoretical ACFs are well within the IQR of the estimated ACFs. 

For $\phi = 0.5$ the estimated IQR of the CBB (green ribbon) is almost identical to the theoretical IRQ (blue ribbon). However, for $\phi = 0.8$ and $\phi = 0.9$ the upper limit of the estimated IQRs for CBB seem to be considerably downward biased for all lags. Noticably, the estimated IQSs for CB (red ribbon) appear to be much less sensitive to the value of $\phi$ and the estimated IQSs for CB are only marginally wider than the theoretical IQRs. We find these results very encouraging.

\begin{figure}[t!]
\centering
\hspace{-1em}\includegraphics[width=\textwidth]{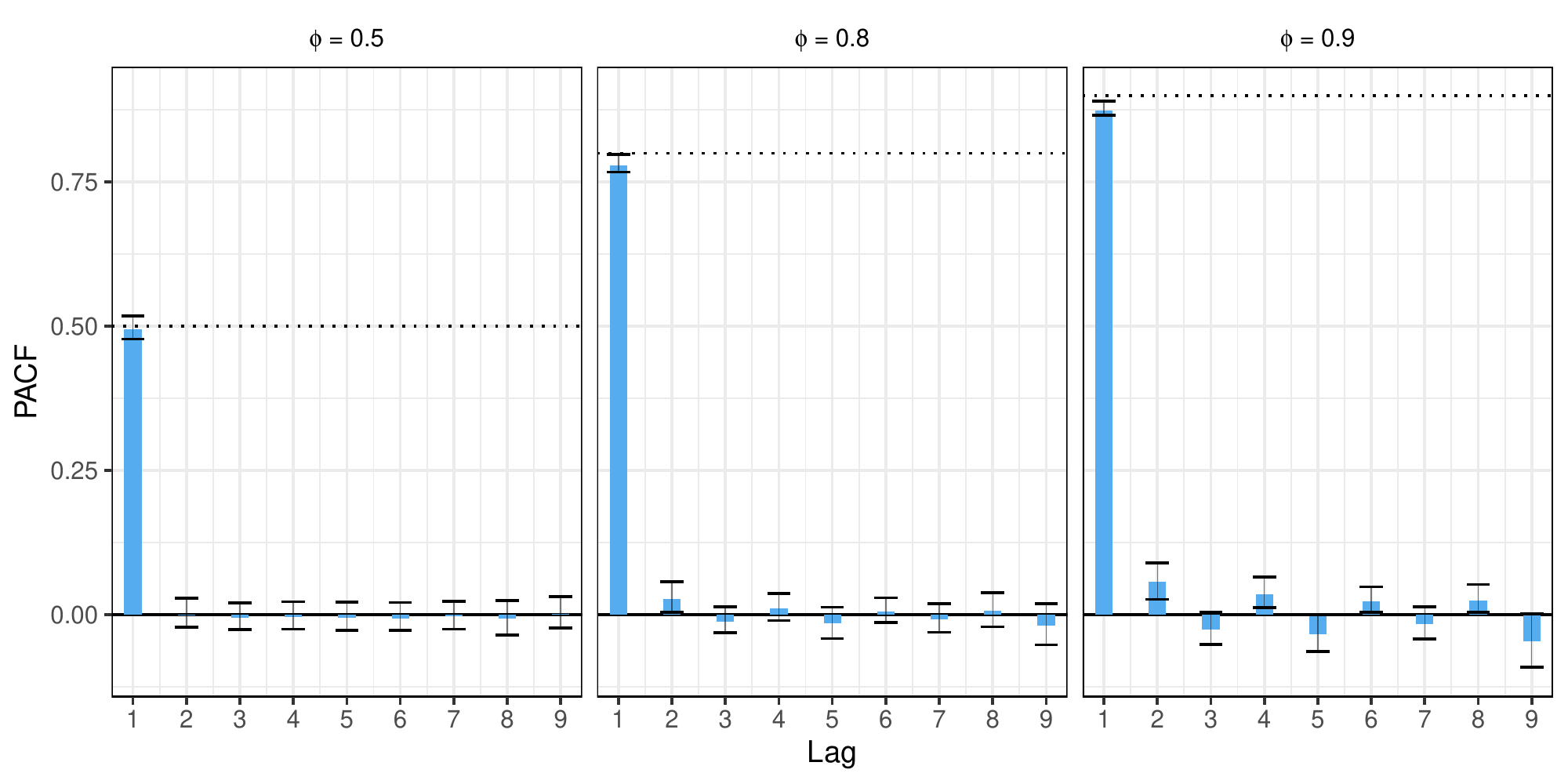}
\captionof{figure}{Top panel: Sample partial autocorrelation function (blue bars) for $\phi = 0.5$, $\phi = 0.8$ and $\phi = 0.9$. The estimates are based on $1,000$ replications of the generative model with $10,000$ samples per replication. The black marks depict the interquartile range (IQR) of the estimates across the $1,000$ replications.}
\label{figure:pacf}
\end{figure}

Next we turn to the partial autocorrelation function (PACF). 
For an AR(p) process the PACF is zero for lags larger than $p$. The AR(1) process is expected to have PACF equal to $\phi$ at lag 1 and zero PACF for all following lags. Figure~\ref{figure:pacf} depicts the estimated PACF using the GAN samples and plots it against the theoretical PACF (horizontal dotted line). The black horizontal marks denote the estimated IQRs. The estimated PACFs have the expected behaviour for $\phi = 0.5, 0.8$ with values close to 0.5 and 0.8 at lag 1 respectively and with values very close to zero for all remaining higher order lags. For the highly persistent case $\phi = 0.9$ the estimated PACF is slightly more imprecise with a notable non-zero PACF at lag 2. However, overall, the PACF based on the GAN sample clearly suggests that the underlying time series under consideration is a highly persistent AR(1) process.

\begin{figure}[t!]
\begin{center}
\includegraphics[width=0.95\textwidth]{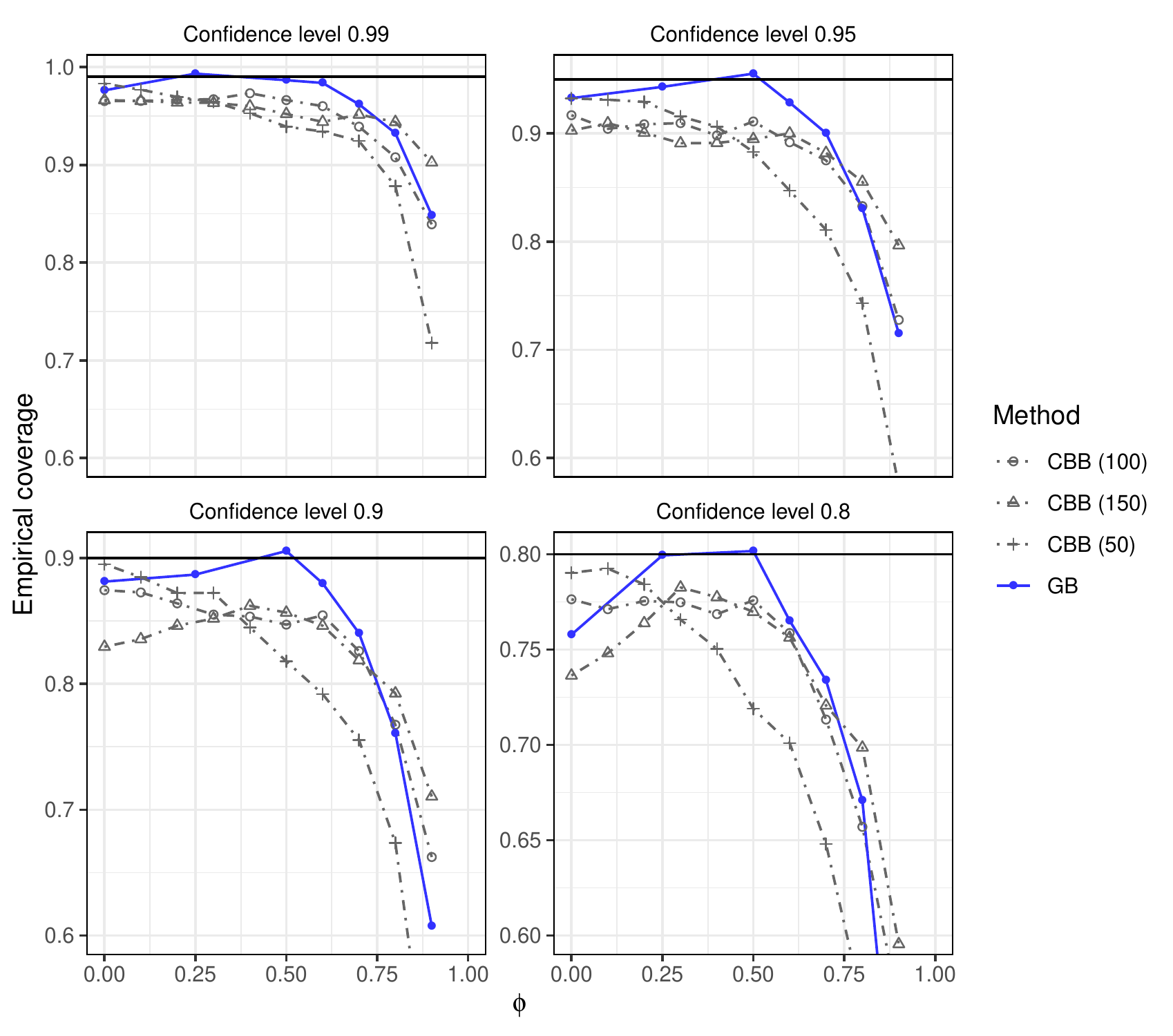}
\captionof{figure}{Empirical coverage of percentile confidence intervals -- with nominal confidence levels $(0.99, 0.95, 0.90, 0.80)$ -- for the CBB and the GB (using complete sampling) under different choices of the autoregressive parameter $\phi$. The horizontal lines depict the corresponding desired nominal confidence levels.} 
\label{fig:blocksizes}
\end{center}
\end{figure}

\textbf{(b) Bootstrapping the least-squares estimator}. We apply the GAN for re-sampling and examine if it recovers higher-level statistics, in particular, the sampling distribution of the usual least-squares (LS) estimator $\hat\phi_{LS}$ of the autoregressive parameter $\phi$. 

The simulation design is identical to that initially described. We obtain $10,000$ sample paths from the GAN across $1,000$ replications with $\phi = 0.5, 0.8, 0.9$. For each replication, the GB variance of $\hat{\phi}_{LS}$ and the GB confidence intervals are constructed as in Section~\ref{sec:genbootstrap}. Using the usual asymptotic approximation, the LS estimator $\hat{\phi}_{LS}$ has asymptotic variance $(1-\phi^2)^{-1}$. 

Figure~\ref{fig:blocksizes} contains the main simulation results for the GB using complete sampling ($b_2 = 1,000$). For values of $\phi$ in the range $0-0.75$ the GB in general produces far better empirical coverage than CBB for all nominal confidence levels, but in particularly for the levels 0.99, 0.95, and 0.9. For the highly persistent processes none of the re-sampling methods considered have good empirical coverage. In these cases the empirical coverage of the GB is at par with the CBB with a block size equal to 100. Not surprisingly, the CBB with the largest block size (=150) here has the best coverage.   

It is very likely that the performance of the GB for highly persistent processes could be improved by increasing the number of layers in temporal convolution network. Recall, that a temporal convolution network with $d$ layers and fixed kernel size $2$ accounts for at-most $2^{d+1}-1$ lags (the receptive field), hence an increase in $d$ might rectify this problem . How to select the optimal number of $d$ layers in the GB procedure as a function of the persistence of original process is ongoing work.

\section{Concluding remarks}\label{sec:conclusion}
GANs provide a promising approach for simulating time series data. Our results suggest that GANs can accurately learn the dynamics of common autoregressive time series processes using temporal convolutional networks. In addition, it seems compelling that the GAN appears to improve empirical coverage in bootstrapping of dependent data when compared to the circular block bootstrap. This gives credibility to the use of GANs on data from time series processes that are unknown.

It is important to note that the various dependent bootstraps have a theoretical justification and their properties have been theoretically derived, see e.g. the overview in \citep{lahiri1999theoretical}. The GAN and GB currently do not have this theoretical reassurance.

The GAN relies on a large number of hyper parameters and design choices. We have not investigated how sensitive our results are to these but research on this is ongoing. We have used sensible defaults for batch sizes, learning rates, gradient penalty, update iteration scheme, activation functions, optimiser but these are by no means optimal choices. This is a general shortcoming in the GAN literature and little is known about how to optimally choose these values.

Our simulations also rely on a simple and basic time series process. It would be fruitful to consider the performance on more general stationary processes (ARMA) and in settings where the error term has stochastic variance, e.g. GARCH.

\newpage
\singlespacing
\printbibliography
\newpage
\doublespacing
\section*{Appendix}
\begin{table}[h!]
\centering
\setlength{\tabcolsep}{0.45em} 
\begin{tabular}{rrrrrrrrrrr}
\toprule\toprule
Hyper parameter & Value \\ \midrule
Discriminator learning rate, $lr_{d}$ & 0.00025 \\
Generator learning rate, $lr_{d}$ & 0.00025 \\
Gradient penalty, $\lambda$ & 20 \\
Batch size, $n_b$ & 64 \\
Discriminator init updates, $N_{init}$ & 50 \\
Discriminator updates, $N_{discriminator}$ & 5 \\
Generator updates, $N_{generator} $& 1 \\
Initial weight distribution & $\text{N}(0, 0.02)$ \\
Adam optimiser, $\epsilon$ & $10^{-8}$ \\
Adam optimiser, $\beta_1$ & 0.5 \\
Adam optimiser, $\beta_2$ & 0.9 \\ \bottomrule\bottomrule
\end{tabular}
\caption{Generative Bootstrap (GB) hyper paramaters.}
\label{table:hypers}
\end{table}

\typeout{get arXiv to do 4 passes: Label(s) may have changed. Rerun}
\end{document}